\documentclass{llncs}

\usepackage{url}
\usepackage{graphicx}

\title{Predicting NCAAB match outcomes using ML techniques -- some results and lessons learned}
\author{Zifan Shi, Sruthi Moorthy, Albrecht Zimmermann\thanks{albrecht.zimmermann@cs.kuleuven.be}}
\institute{KU Leuven, Belgium}

\begin{document}

\maketitle

\begin{abstract}
Most existing work on predicting NCAAB matches has been developed in a statistical context. Trusting the capabilities of ML techniques, particularly classification learners, to uncover the importance of features and learn their relationships, we evaluated a number of different paradigms on this task. In this paper, we summarize our work, pointing out that attributes seem to be more important than models, and that there seems to be an upper limit to predictive quality.
\end{abstract}

\section{Introduction}

Predicting the outcome of contests in organized sports can be attractive for a number of reasons such as betting on those outcomes, whether in organized sports betting or informally with colleagues and friends, or simply to stimulate conversations about who ``should have won''. We would assume that this task is easier in professional leagues, such as Major League Baseball (MLB), the National Basketball Association (NBA), or the National Football Association (NFL), since there are only relatively few teams and their quality does not vary too widely. As an effect of this, match statistics should be meaningful early on since the competition is strong, and teams play the same opponents frequently. Additionally, professional leagues typically play more matches per team per season, e.g. 82 in the NBA or 162 in MLB, than in college or amateur leagues in which the sport in question is (supposed to be) only a side aspect of athletes' lives.

National College Athletics Association Basketball (NCAAB) matches therefore offer a challenging setting for predictive learning: more than 300 teams that have strongly diverging resource bases in terms of money, facilities, and national exposure and therefore attractiveness for high quality players, play about 30 games each per season, can choose many of their opponents themselves (another difference to professional teams), and often have little consistency in the composition of teams from one season to the next since especially star players will quickly move on to professional sports. Lopsided results and unrealistic match statistics will therefore not be uncommon, distorting the perception of teams' quality.

Most of the existing work in the field is more or less statistical in nature, with much of the work developed in blog posts or web columns. Many problems that can be addressed by statistical methods also offer themselves up as Machine Learning settings, with the expected advantage that the burden of specifying the particulars of the model shifts from a statistician to the algorithm. Yet so far there is relatively little such work in the ML literature. The main goal of the work reported in this paper was therefore to assess the usefulness of classifier learning for the purpose of predicting the outcome of individual NCAAB matches. Several results of this work were somewhat unexpected to us:
\begin{itemize}
\item Multi-layer perceptrons, an ML technique that is currently not seeing wide-spread use, proved to be most effective in the explored settings.
\item Explicitly modeling the differences between teams' attributes \emph{does not} improve predictive accuracy. 
\item Most interestingly, there seems to be a ``glass ceiling'' of about 74\% predictive accuracy that cannot be exceeded by ML or statistical techniques.
\end{itemize}


\section{\label{definitions}Definitions}

The most straight-forward way of describing basketball teams in such a way that success in a match can be predicted relate to scoring points -- either scoring points offensively or preventing the opponent's scoring defensively. Relatively easy to measure offensive statistics include field goals made (FGM), three-point shots made (3FGM), free throws after fouls (FT), offensive rebounds that provide an additional attempt at scoring (OR), but also turnovers that deprive a team of an opportunity to score (TO). Defensively speaking, there are defensive rebounds that end the opponent's possession and give a team control of the ball (DR), steals that have the same effect and make up part of the opponent's turnovers (STL), and blocks, which prevent the opponent from scoring (BLK). And of course, there are points per game (PPG) and points allowed per game (PAG).

The problem with these statistics is that they are all raw numbers, which limits their expressiveness. If a team collects 30 rebounds in total during a game, we cannot know whether to consider this a good result unless we know how many rebounds were there to be had in the first place. 30 of 40 is obviously a better rebound rate than 30 of 60. Similar statements can be made for field goals and free throws, which is why statistics like offensive rebound rate (ORR), turnover rate (TOR), or field goals attempted (FGA) will paint a better picture. Even in that case, however, such statistics are not normalized: 40 rebounds in a game in which both teams combined to shoot 100 times at the basket is different from 40 rebounds when there were only 80 scoring attempts.

For normalization, one can calculate the number of possessions in a given game:
$$Possessions = 0.96*(FGA - OR - TO + (0.475*FTA))$$
and normalize teams' points scored and allowed per 100 possessions, deriving offensive and defensive \emph{efficiencies}:
$$OE = \frac{Points\ scored * 100}{Possessions}, DE = \frac{Points\ allowed * 100}{Possessions}$$
It should be noted that the factor $0.475$ is empirically estimated -- when first introducing the above formulation for the NBA, Dean Oliver estimated the factor as $0.4$ \cite{basketball-on-paper}.

Dean Oliver has also singled out four statistics as being of particular relevance for a team's success, the so-called ``Four Factors'' (in order of importance, with their relative weight in parentheses):
\begin{enumerate}
\item Effective field goal percentage (0.4):
$$eFG\% = \frac{FGM + 0.5 \cdot 3FGM}{FGA}$$
\item Turnover percentage (0.25):
$$TO\% = \frac{TO}{Possessions}$$
\item Offensive Rebound Percentage (0.2):
$$OR\% = \frac{OR}{(OR + DR_{Opponent})}$$
\item Free throw rate (0.15):
$$FTR = \frac{FTA}{FGA}$$
\end{enumerate}

While such statistics are normalized w.r.t. the ``pace'' of a game, they do not take the opponent's quality into account, which can be of particular importance in the college game: a team that puts up impressive offensive statistics against (an) opponent(s) that is (are) weak defensively, should be considered less good than a team that can deliver similar statistics against better-defending opponents. For best expected performance, one should therefore normalize w.r.t. pace, opponent's level, and national average, deriving \emph{adjusted} efficiencies:
$$AdjOE = \frac{OE * avg_{all\ teams}(OE)}{AdjDE_{opponent}}, AdjDE = \frac{DE * avg_{all\ teams}(DE)}{AdjOE_{opponent}}$$

To gain a comprehensive picture of a team's performance during the season, such statistics would have to be averaged over all games (we describe two approaches for doing so in Section \ref{subsec:adjusted-effs}), and a state-of-the-art way of using the derived statistics in predicting match outcomes consists of using the so-called Pythagorean Expectation, e.g.:
$$Win\ Probability = \frac{((Adjusted)\ OE_{avg})^y}{((Adjusted)\ OE_{avg})^y + ((Adjusted)\ DE_{avg})^y}$$
to calculate each team's win probability and predicting that the team with the higher probability wins. More generally, \emph{ranking systems} can by used by ranking the entire pool of teams and predicting for each match-up that the higher ranked team wins.

\section{\label{related}Related Work}

The use of the Pythagorean Expectation actually goes back to Bill James' work on baseball. It was adapted for the use in basketball prediction by numerous analysts, including such luminaries as Daryl Morey, John Hollinger, Ken Pomeroy, and Dean Oliver. The difference between the different approaches comes down to which measures of offensive and defensive prowess are used and how the exponent has been estimated. Dean Oliver was also the one who first introduced possession-based analysis formally in his book ``Basketball on Paper'' \cite{basketball-on-paper}, although he acknowledges that he had seen different coaches use such analysis in practice. In the same work, he introduced the ``Four Factors''.

The adjustment of efficiencies to the opponent's quality is due to Ken Pomeroy who uses them as input in his version of the Pythagorean Expectation to rank NCAAB teams and predict match outcomes. His is far from the only ranking system, however, with other analysts like Jeff Sagarin, Ken Massey or Raymond Cheung running their own web sites and giving their own predictions. Comparisons of the results of different ranking systems can for instance be found at \url{http://masseyratings.com/cb/compare.htm}  or \url{http://www.raymondcheong.com/rankings/perf13.html}. The worst accuracy for those systems is in the $62\%-64\%$ range, equivalent to predicting that the home team wins, the best ones achieve up to $74\%-75\%$.

The NCAAB itself uses the so-called Ratings Percentage Index to rank teams, a linear weighted sum of a team's winning percentage, its opponents' winning percentage, and the winning percentage of those opponents' opponents.

As an alternative approach, Kvam \emph{et al.} have proposed a logistic regression/Markov chain model \cite{journal/nrl/kvam2006}. In this method, each team is represented as a state in a Markov chain and state transitions occur if one team is considered better than its opponent. Logistic regression is used to estimate transition probability parameters from the data. The authors have proposed an updated version using Bayesian estimates \cite{journal/jqas/brown2010}, and recently published work in which they estimate their method's success in comparison to other ranking schemes \cite{conf:sloans/2012/brown}.

\section{\label{predictions}Day-by-day predictions using ML}

The approaches described in the preceding section are in many cases somewhat or even fully hand-crafted. This can be rather high-level, as in \emph{defining} the transition probabilities in LRMC's Markov chain by hand, or it can go as far as Ken Pomeroy taking home court advantage into consideration by \emph{multiplying} the home team's stats by $1.014$.
Furthermore, especially the Pythagorean Expectation seems to be a rather simple model. 

Machine Learning promises to address both of these issues: we would expect to be able to \emph{learn} the relative importance of different descriptive measures, in particular if this importance changes for different numerical ranges, and to be able to \emph{learn} their relationships, automatically making the model as difficult (or simple) as needed. We therefore turned to classification learners representing several different paradigms and evaluated their performance.

In a reversal of current practice, explicit prediction of match outcomes could be used to rank teams by predicting the outcome of all hypothetical pairings and ranking teams by number of predicted wins.

\ \\
\noindent The evaluated learners were:
\begin{itemize}
\item Decision trees, represented by C4.5.
\item Rule learners, represented by Ripper.
\item Artificial neural networks, represented by a Multi-layer Perceptron (MLP).
\item Na{\"i}ve Bayes
\item Ensemble learners, by a random forest.
\end{itemize}
All algorithms were used in the form of their respective Weka implementations and run with default parameter settings, with the exception of Na{\"i}ve Bayes, for which the ``Kernel Estimator''  option was activated to enable it to handle numerical attributes effectively, J48, whose pre-pruning threshold we set to $1\%$ of the training data, and the Random Forest, which we set to consist of $20$ trees.
All data has been downloaded from Ken Pomeroy's web site, \url{kenpom.com}, and we limit ourselves to matches involving two Division I teams. Matches were encoded by location (home, away, neutral court), the chosen numerical statistics up to the day the match was played, and the outcome (win, loss) from the perspective of the first team. We always chose the team with the lexicographically smaller name as first team. For each experiment run, one season was used as test set and the preceding seasons from 2008 onward as training data, leading to the training and test set sizes shown in Table \ref{ds-sizes}.

\begin{table}
\begin{center}
\begin{tabular}{l|c|c|c|c|c}
Season &  2009 &2010 &2011&2012&2013\\\hline
Train & 5265 &10601& 15990&21373&26772\\
Test & 5336 &5389&5383&5399&5464\\
\end{tabular}
\caption{Training and test set sizes per season\label{ds-sizes}}
\end{center}
\end{table}

\subsection{Seasonal Averaging}

Ken Pomeroy's web site features only the most recent averaged adjusted efficiencies (and averaged Four Factors), i.e. from the end of the season for completed seasons, and for seasons in progress the efficiencies up to the current date. We therefore calculated the day-to-day averaged adjusted efficiencies ourselves, following Pomeroy's description. While that description is very precise for the most part, the averaging is summarized as averaging over the season with more weight given to recent games. We chose to average via two methods:

\begin{enumerate} 
\item an adjustable weight parameter $\alpha$:
$$AdjE_{avg,post-match} = (1-\alpha) AdjE_{avg,pre-match} + \alpha AdjE_{post-match}$$
and evaluated a number of different alpha values. Both averaged efficiencies and Four Factors stabilized for $\alpha=0.2$. To have a pre-match value for the first game of the season, we used the preceding season's end-of-season efficiencies, and
\item explicitly:\\
A side-effect of using an $\alpha$-parameter less than $0.5$ (e.g. 0.2) in averaging is that last season's end-of-season averaged adjusted efficiency is weighted rather highly since it is the only value whose weight is never multiplied with $\alpha$ itself but always with $(1-\alpha)$. 
We therefore evaluated a different weighting scheme in which each match's adjusted efficiency is weighted explicitly with the number of games played $+1$. This means that last season's end-of-season efficiency has weight one, the adjusted efficiency of the first game weight two etc. The sum is normalized with the total sum of weights up to the current date. 
\end{enumerate}

We have to admit that using either way, we did not manage to arrive at the same end-of-season efficiencies as Ken Pomeroy. Typically, our values are more extreme, with adjusted offensive efficiencies higher and adjusted defensive efficiencies lower than Pomeroy's values.
Also, since $\alpha$-weighting performed consistently worse, we will focus on the explicit averaging for the rest of the paper.

\subsection{Using adjusted efficiencies\label{subsec:adjusted-effs}}

In the first set of experiments, we aimed to identify which attributes out of the full set of raw statistics, normalized statistics, Four Factors, and adjusted efficiencies were most useful in predicting match outcomes. We found the combinations of location and adjusted offensive and defensive efficiencies, and location and Four Factors to work best. This result is supported by the outcome of using Weka's feature selection methods to winnow the attribute set down, which select location first, followed by adjusted efficiencies, and the Four Factors.

A somewhat surprising result is the weak performance of the symbolic classifiers: MLP and Na{\"i}ve Bayes give consistently best results (Table \ref{adjusted-efficiencies}). We also see that more training data does not translate into better models, and that 2012 seems to have been an outlier season.





\begin{table*}
 \begin{tabular}{c@{\hspace{1cm}}c}
 \begin{minipage}{0.45\textwidth}
\begin{center}
\begin{tabular}{l|c|c|c|c}
Season & J48 & RF & NB & MLP\\\hline
2009 & 0.6839 & 0.6885 & 0.7101 & 0.7077\\
2010 & 0.6899 & 0.6942 & 0.7172 & 0.7251\\
2011 & 0.6905 & 0.6779 & 0.7028 & 0.716\\
2012 & 0.7042 & 0.7137 & 0.7276 & 0.7446\\
2013 & 0.6898 & 0.6881 & 0.7193 & 0.7215\\
\end{tabular}
\caption{Match outcome prediction accuracies using adjusted efficiencies\label{adjusted-efficiencies}}
\end{center}
\end{minipage}
&
\begin{minipage}{0.45\textwidth}
\begin{center}
\begin{tabular}{l|c|c|c|c}
Season & J48 & RF & NB & MLP\\\hline
2009 & 0.6647 & 0.6801 & 0.7121 & 0.7011\\
2010 & 0.6645 & 0.6931 & 0.7202 & 0.7165\\
2011 & 0.6622 & 0.6983 & 0.7206 & 0.7121\\
2012 & 0.6788 & 0.702 & 0.7305 & 0.7311\\
2013 & 0.6508 & 0.6892 & 0.7081 & 0.7092\\
\end{tabular}
\caption{Match outcome prediction accuracies using adjusted four factor\label{explicitly-adjusted-ff}}
\end{center}
\end{minipage}
\end{tabular}
\end{table*}
%
%

The accuracies for the different seasons are on par with those of the best-performing predictive systems, e.g. Ken Pomeroy's predictions and the LRMC, but unfortunately they are not better. 

\subsection{Using adjusted Four Factors}

As mentioned in Section \ref{related}, Dean Oliver proposed the so-called ``Four Factors'' as being influential for a team's success. Since our experiments had indicated that the unadjusted Four Factors were already as useful in predicting match outcomes as adjusted efficiencies, we assumed that adjusted Four Factors should be more effective. We therefore performed adjusting in the same way as for efficiencies: multiplying with the national average and dividing by the opponent's counter-statistic, averaging using both methods. Averaging using $\alpha$ proved again to be worse, while explicitly averaging lead to similar yet slightly worse results compared to using adjusted efficiencies, as Table \ref{explicitly-adjusted-ff} shows. 

In a bid to improve the performance of the symbolic classifiers, we also experimented with encoding the differences between adjusted Four Factors explicitly, hypothesizing that for instance C4.5's over-fitting had to do with inducing branches for many different combinations of values that could be summarized by their difference. We either subtracted a team's defensive factor from the opponent's corresponding offensive factor, or subtracted offensive from corresponding offensive, and defensive from corresponding defensive factors. The former scheme severely underperformed, while the latter scheme with explicit weights for averaging showed very similar results to Table \ref{explicitly-adjusted-ff}.

Finally, we attempted to address our more extreme adjusted values by calculating each season from stretch, not using the preceding season's values as input for the first game. While the resulting adjusted efficiencies are closer to those reported on Pomeroy's web site, prediction accuracies also decrease slightly.

\subsection{Development of predictive accuracy as the season progresses}

\begin{figure}
\begin{center}
\includegraphics[angle=270,width=0.8\textwidth]{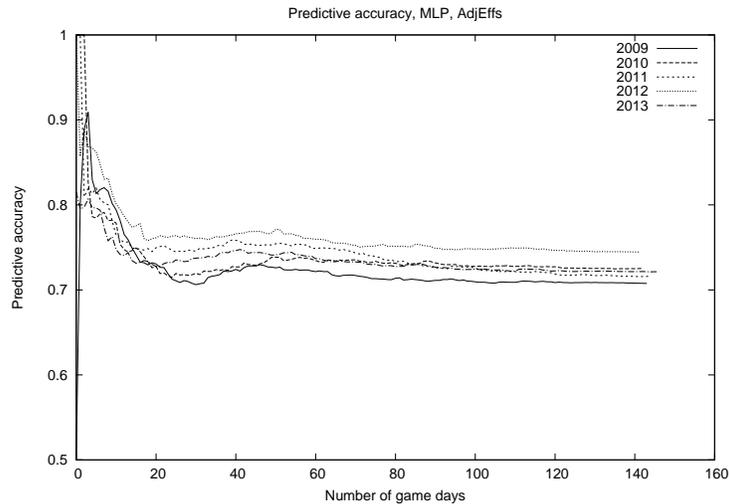}
\caption{Development of predictive accuracy over the course of a season (MLP, AdjEff)\label{season-curve-adjeff-mlp}}
\end{center}
\end{figure}

Figure \ref{season-curve-adjeff-mlp} shows how predictive accuracy develop as the season progresses. We chose MLP with adjusted efficiencies for this plot but the general trend is representative for other settings.

With the exception for 2009, when only training data from 2008 was available, predictive accuracy is $100\%$ or close to it for the first few days of the season and then experiences a dip before it recovers, and shows only slight deterioration for the rest of the season. Interesting, but hard to spot in the plot, is that there are small up-and-downs in the playoffs, particularly in the last rounds, for instance predicting the semi-finals and final correctly after getting the quarter-finals wrong. 

%
\section{\label{conclusions}Lessons learned and open questions}

In this work, we have explored the use of ML techniques, specifically classification learners, for making NCAAB match outcome predictions. These are just preliminary steps and the exploration is obviously far from complete. While the results were somewhat disappointing, we want to stress that they were not bad per se -- being on par with the state-of-the-art is only disappointing since we aimed to improve on it. Given our results, however, we believe that there are two first lessons that can be learned and that should guide our next steps.

\subsection{It's in the attributes, not in the models}

As stated above, one of our expectations was that more complex models could tease out relationships that simpler models would miss. Instead, we found that Na{\"i}ve Bayes, arguably the simplest of the classifiers, performs remarkably well. Similar observations can actually be made about existing techniques, since Ken Pomeroy's straight-forward Pythagorean Expectation performs as well as, or even better than, the much more complex LRMC model, Brown \emph{et al.}'s claims notwithstanding.

Instead, whatever differences in performance we have observed essentially came down to the used attributes and how they were calculated: adjusted efficiencies and (adjusted) Four Factors are validated both by feature selection techniques and by the success of the classifiers trained on those representations but different ways of averaging over the season have an effect on the quality. Using other or additional features, on the other hand, leads to worse results. In a sense, this should not be surprising: any given match will be won by the team that scored more points than the other one, which is the information encoded in the adjusted efficiencies, for instance.

Of course there is also ML/DM conventional wisdom that the main aspect of using such techniques effectively consists of constructing the right representation. Still, we found it surprising how stark the influence of choosing the right attributes was on achieving best results. 

\subsection{There seems to be a glass ceiling}

Which brings us to the second lesson: the other invariant that we saw in our experiments is that there seems to be an upper limit to predictive accuracy for match outcomes, at around $74\%-75\%$. This holds not only for Na{\"i}ve Bayes and the MLP, but when one considers comparisons of non-ML methods, e.g. \url{http://www.raymondcheong.com/rankings/perf13.html} or \cite{conf:sloans/2012/brown}, one finds similar results. Additionally there are works in fields such a soccer \cite{DBLP:conf/ijcnn/HuangC10} (76.9\%), American Football \cite{535226} (78.6\%), NCAA Football \cite{pardee1999artificial} (76.2\%), and the NBA \cite{RePEc:bpj:jqsprt:v:5:y:2009:i:1:n:7} (74.33\%) that show best results in a similar region.

It is difficult to determine why this is the case. If the claim made in the preceding section holds and the performance of predictors comes down to attribute construction, then maybe this glass ceiling is an artifact of the attributes we and others use. It is also possible, however, that there is simply a relatively large residue of college basketball matches that is in the truest sense of the world unpredictable.

\subsection{Where to next?}

First off, there is need to verify that our first lesson is correct and attributes are indeed what make or break success. To this end, different feature selection and modeling techniques need to be contrasted to get a clear understanding of attributes' effects, and how to best aggregate them over the course of a season. Following (or parallel to) that, both of the possible explanations for the glass ceiling given above offer themselves up for exploration that we intend to pursue in the near future:

1) Most existing attributes do not encode so-called ``intangibles'' such as experience, leadership, or luck. Attempts have been made to construct objective indicators, as in \url{http://harvardsportsanalysis.wordpress.com/2012/03/14/survival-of-the-fittest-a-new-model-for-ncaa-tournament-prediction/}, whose author proposes a ``Returning Minutes Percentage'', Dean Oliver's attempts to measure positional stability, or Ken Pomeroy's work that takes the luck of teams into account. Pomeroy incidentally credits Dean Oliver (once again) with having introduced this into basketball analysis. Hence, constructing new attributes that include additional information could improve predictive power. 

2) A better understanding of incorrectly predicted matches is necessary. The weak performance of ensembles indicates that misclassified matches are not easily modeled. However, identifying similarities of misclassified matches or learning a model that can discriminate correctly and incorrectly classified instances, would help in  gaining an understanding whether those matches are different or simply unpredictable. At this point, we would also finally come back to whether we can determine which team ``should have won''. 

Finally, somewhat unrelated, it could be interesting to separate training data by conference and learn models particular to the involvement of certain conferences teams. The most challenging question would probably have to do with how to decide which model's prediction to use if the two models disagree.

\bibliographystyle{plain}
\bibliography{../bibliographie}

\end{document}